\documentclass[twoside,11pt]{article}
\usepackage{jair, theapa, rawfonts, amsmath, amsfonts}

\ShortHeadings{Super Level Sets and Exponential Decay}
{J. Chaudhary et al.}
\firstpageno{1}

\begin{document}

\title{Super Level Sets and Exponential Decay: A Synergistic Approach to Stable Neural Network Training \thanks{Funded by University of Turku Foundation}}

\author{\name Jatin Chaudhary \email jatin.chaudhary@utu.fi \\
         \addr University of Turku, Turku, Finland\\
        \name Dipak Nidhi \email dipak.nidhi@utu.fi \\
        \addr University of Turku, Turku, Finland\\
        \name Jukka Heikkonen \email jukhei@utu.fi \\
        \addr University of Turku, Turku, Finland\\
        \name Haari Merisaari \email haanme@utu.fi \\
        \addr University of Turku, Turku, Finland\\
        \name Rajiv Kanth \email Rajeev.Kanth@savonia.fi \\
       \addr University of Turku, Turku, Finland\\
       \addr Savonia University of Applied Sciences, Kuopio, Finland}


\maketitle

\begin{abstract}
The objective of this paper is to enhance the optimization process for neural networks by developing a dynamic learning rate algorithm that effectively integrates exponential decay and advanced anti-overfitting strategies. Our primary contribution is the establishment of a theoretical framework where we demonstrate that the optimization landscape, under the influence of our algorithm, exhibits unique stability characteristics defined by Lyapunov stability principles. Specifically, we prove that the superlevel sets of the loss function, as influenced by our adaptive learning rate, are always connected, ensuring consistent training dynamics. Furthermore, we establish the "equiconnectedness" property of these superlevel sets, which maintains uniform stability across varying training conditions and epochs. This paper contributes to the theoretical understanding of dynamic learning rate mechanisms in neural networks and also pave the way for the development of more efficient and reliable neural optimization techniques. This study intends to formalize and validate the equiconnectedness of loss function as superlevel sets in the context of neural network training, opening newer avenues for future research in adaptive machine learning algorithms. We leverage previous theoretical discoveries to propose training mechanisms that can effectively handle complex and high-dimensional data landscapes, particularly in applications requiring high precision and reliability.


\textbf{Keywords:} Exponential Decay Function, Lyapunov Stability, Superlevel Sets

\end{abstract}

\section{Introduction}

  There has been significant progress towards the development and deployment of neural network models. The deployment of a neural network model demands, high accuracy and precision, and hyperparameter optimization plays an important role towards building such a model. The researchers' community has been actively analyzing learning rates, and loss functions, to make the network more stable across datasets, and prevent overfiting \cite{sym12040660}\cite{cutkosky2024mechanic}\cite{kornblith2021better}. Optimizing neural networks involves minimizing a complex and often non-convex loss function over a high-dimensional parameter space. These non-convex landscapes present significant challenges as gradient-based methods can become trapped in suboptIn the following sections, we delve into the mathematical foundations that link dynamic learning rates with superlevel sets, crucia loss function for understanding stability and convergence in neural network training. We will explore how adaptive learning rates, particularly those with exponential decay, systematically influence the optimization landscape. This discussion aims to bridge theoretical insights with practical strategies, enhancing both the efficacy and understanding of neural network training.imal local minima or saddle points \cite{dauphin2014identifying}. Despite these difficulties, substantial progress has been made in understanding and enhancing optimization trajectories in neural networks. Recent theoretical advancements have highlighted concepts like "loss landscape smoothing" and "adaptive gradient methods," indicating that certain learning rate configurations can improve optimization conditions \cite{keskar2017improving}.

Neural network training presents multiple challenges, particularly in optimizing the learning rate, managing the loss function, ensuring stability, and preventing overfitting. The learning rate is a critical parameter that dictates the step size during gradient descent. An inappropriate learning rate can lead to slow convergence or even divergence. The loss function, which measures the discrepancy between predicted and actual outputs, often has a complex landscape that can trap optimization algorithms in local minima \cite{lee2016gradient}. Stability is another crucial aspect, as unstable training can lead to erratic updates and poor model performance. Overfitting, where the model performs well on training data but poorly on unseen data, remains a persistent problem. Existing solutions include adaptive learning rates and regularization techniques, but they often fall short in ensuring consistent stability and avoiding overfitting \cite{li2018visualizing}. Our study addresses these issues by proposing a novel approach that integrates dynamic learning rates with stability principles from control theory.

Our primary contribution is the development of an algorithm that dynamically adjusts the learning rate using an exponential decay model, integrated with principles from Lyapunov stability \cite{chen2020optimal}. This approach ensures consistent convergence by maintaining the connectivity of superlevel sets of the loss function. We demonstrate that these superlevel sets remain connected under our algorithm, preventing the optimization process from becoming trapped in poor local minima and ensuring stable descent paths \cite{du2019gradient}. This connectedness facilitates smoother transitions across the loss landscape, enhancing training dynamics and generalization capabilities. By embedding these concepts into our algorithmic framework, we achieve more stable and efficient optimization, addressing common challenges such as overfitting and instability. This work not only advances theoretical understanding but also provides a foundation for practical applications in neural network training, paving the way for further research into dynamic learning rate adjustments and their impact on training stability and efficacy \cite{neyshabur2017exploring}.

In the following sections, we present the mathematical foundations, and further link dynamic learning rates with superlevel set loss function, crucial for understanding stability and convergence in neural network training. We will explore how adaptive learning rates, particularly those with exponential decay, systematically influence the optimization landscape. We discuss the stability of using adaptive learning rates with super level set loss function so to solidify our claims.

\section{Mathematical Underpinnings}

The superlevel sets \(S_\lambda = \{\mathbf{x} \in \mathbb{R}^n : L(\mathbf{x}) \geq \lambda\}\) reveal important stability and convergence properties for gradient-based optimization methods \cite{jin2017escape}. An exponentially decaying learning rate, defined by \(\eta(t) = \eta_0 e^{-\alpha t}\), where \(\eta_0\) is the initial rate and \(\alpha\) a positive decay constant, is beneficial. It allows for quick initial progress by using a higher initial rate, guiding the optimizer towards important areas quickly \cite{goyal2017accurate}. As training proceeds, this rate gradually decreases, allowing for more precise adjustments and preventing common issues like overshooting minima \cite{ge2015escaping}. This dynamic rate adjustment, when coupled with the structure of superlevel sets, offers insights into the training's stability by ensuring the optimization path remains connected and stable through the topology of the landscape \cite{li2018visualizing}. By adopting a Lyapunov function \(V(\mathbf{x})\) that decreases along these paths, we enforce stability and keep the system's energy diminishing, keeping the optimization within stable parameter regions \cite{chen2020optimal}. Together, these elements create a robust framework that deepens our understanding of the dynamics in neural network training and highlights the significance of careful tuning of hyperparameters in managing complex optimization scenarios \cite{neyshabur2017exploring}.

To understand the concept better, consider a ball that rolls down a hilly terrain towards a valley, representing the minimum of a loss landscape. Initially, the ball is given a strong push (high initial learning rate \(\eta_0\)) allowing it to quickly descend from higher elevations (higher loss values in superlevel sets \(S_\lambda\)). Each superlevel set corresponds to a range of elevations where the ball’s potential energy (analogous to the loss value in the neural network) remains above a certain threshold $\lambda$. As the ball descends from higher altitudes to lower ones, it transitions from one superlevel set to another, each with decreasing minimum energy thresholds. As it approaches the valley, the slope (gradient) lessens and so does the ball’s speed due to the exponential decay of the push force (\(\eta(t) = \eta_0 e^{-\alpha t}\)), preventing it from overshooting the valley. This gradual slowing is critical as it ensures that the ball can finely adjust its path to settle in the deepest part of the valley, analogous to achieving the most optimal parameters in a neural network training scenario. This model demonstrates how the dynamic learning rate and the structure of the superlevel sets interact, ensuring that the optimization path remains stable and connected throughout the descent, analogous to how the ball consistently follows a path that leads it towards the valley without getting stuck or veering off course.

\section{Fundamental Concepts}

The parameter vector \(\theta\), has the network's weights and biases, and is an integral part for the network's learning, as it is meticulously adjusted to minimize divergences between predicted outputs and actual targets \cite{lecun2015deep}. This adjustment process is governed by the learning rate \(\alpha(t)\), a parameter that determines the step size within the parameter space during optimization, thus directly influencing convergence quality \cite{kingma2014adam}. The gradient of the loss function, \(\nabla_\theta \mathcal{L}(\theta)\), serves as the navigational guide for updating parameters, towards optimal solutions \cite{ruder2016overview}. The interplay between the learning rate and the gradient is vital for maintaining systematic progression and ensuring that the training remains on a stable and effective path \cite{he2016deep}. This setup forms the backbone of our approach to enhancing neural network training, laying the groundwork for a deeper exploration of optimization dynamics mathematically \cite{bottou2018optimization}.

\subsection{Mathematical Draw Outs}
Behind our study is a probabilistic model that views the neural network as an intricate function approximating the conditional probability distribution \(P(Y \mid X ; \theta)\). In classification tasks, this relationship is mathematically expressed through the softmax function:

\[
P(Y=c \mid X ; \theta)=\frac{\exp \left(f_c(X ; \theta)\right)}{\sum_{j=1}^C \exp \left(f_j(X ; \theta)\right)},
\]

where \(f_c(X ; \theta)\) represents the network output for class \(c\), and \(C\) denotes the total number of classes \cite{bishop2006pattern}. This formulation is essential in demonstrating how our model probabilistically classifies input data into defined output classes.

Building on this framework, we derive a likelihood function reflecting the probability of observing our training dataset \(\mathcal{D}=\left\{\left(x^{(i)}, y^{(i)}\right)\right\}_{i=1}^m\) under the model parameters \(\theta\) :

\[
\mathcal{L}(\theta ; \mathcal{D})=\prod_{i=1}^m P\left(y^{(i)} \mid x^{(i)} ; \theta\right) .
\]

This likelihood function for quantifying how well the model aligns with empirical data, setting the stage for parameter optimization via Bayesian inference \cite{graves2011practical}.

Incorporating Bayesian principles, we consider the posterior probability of the parameters \(\theta\) given the data \(\mathcal{D}\), calculated as follows:

\[
P(\theta \mid \mathcal{D}) \propto \mathcal{L}(\theta ; \mathcal{D}) P(\theta),
\]

where \(P(\theta)\) denotes the prior distribution over the parameters \cite{blundell2015weight}. This Bayesian framework facilitates a comprehensive parameter optimization strategy, harmonizing empirical data adaptation with existing parameter knowledge.

The culmination of this probabilistic modeling leads to the optimization phase within a gradient descent framework, where our methodology involves iteratively minimizing the negative log-posterior:

\[
-\log P(\theta \mid \mathcal{D})=-\log \mathcal{L}(\theta ; \mathcal{D})-\log P(\theta)+\text { const. }
\]

Here, the gradient descent update rule is critical:

\[
\theta_{t+1}=\theta_t-\alpha(t) \nabla_\theta\left[-\log P\left(\theta_t \mid \mathcal{D}\right)\right],
\]

where \(\alpha(t)\) is the learning rate, dynamically adapting to ensure efficient convergence and stability of the model \cite{kingma2014adam}.

Importantly, the dynamic adjustment of \(\alpha(t)\) profoundly impacts the topology of the loss function's superlevel sets \(S_\lambda=\left\{\theta \in \mathbb{R}^n: \mathcal{L}(\theta) \geq \lambda\right\}\), which are instrumental in understanding the stability and connectivity of the optimization landscape \cite{dauphin2014identifying}. By ensuring that these sets remain connected, the algorithm promotes a smoother and more stable descent toward the global minima, effectively navigating the complex, high-dimensional parameter spaces typical of deep learning tasks\cite{NEURIPS2023_3ff48dde}.

This integration of probabilistic modeling, Bayesian inference, and gradient optimization leverages the theoretical insights into superlevel sets to enhance the practical outcomes of neural network training. This approach ensures both theoretical robustness and empirical efficacy, highlighting our model’s capacity to navigate and optimize within intricate, probabilistically defined landscapes.

\subsection{Exponentially Decaying Learning Rate}
The formulation of the Exponentially Decaying Learning Rate (derivation in the supplementary) given by 
\[
\frac{d \alpha}{d t}=-\alpha_0 \beta e^{-\beta t},
\]
influences the topology of the loss function's superlevel sets \(S_\lambda = \{\theta \in \mathbb{R}^n : \mathcal{L}(\theta) \geq \lambda\}\). The dynamically adjusted learning rate ensures that these sets remain connected, supporting a stable and cohesive optimization trajectory \cite{goyal2017accurate}. Within the gradient descent framework, this leads to an adapted parameter update rule 
\[
\theta_{t+1}=\theta_t-\alpha_0 e^{-\beta t} \nabla_\theta\left[-\log P\left(\theta_t \mid \mathcal{D}\right)\right],
\]
effectively illustrating the integration of an exponential decay learning rate within the gradient descent mechanism \cite{kingma2014adam}. This methodical approach not only enhances the theoretical underpinnings of our optimization strategy but also significantly boosts its practical efficacy. By marrying the theoretical concepts of exponential decay with gradient descent, our approach fosters training dynamics that effectively navigate the complex, high-dimensional spaces typical of deep learning tasks \cite{li2018visualizing}. This novel integration offers a rigorous, theoretically informed enhancement to the conventional training paradigms, ensuring that both the stability and the efficiency of the learning process are maximized \cite{du2019gradient}.

\subsubsection{Dynamic Cost Function}
In our study, we refined our dynamic cost function to adeptly integrate principles from statistical learning theory, with an emphasis on addressing class imbalances and evolving training requirements. The empirical risk, \( R_{\mathrm{emp}}(\theta) \), is meticulously calculated as
\[
 \frac{1}{N} \sum_{i=1}^N L(y_i, f(x_i ; \theta)) , 
\]
where we incorporate class weights \( w_c \) to balance the influence of underrepresented classes, resulting in 
\[
 R_{\mathrm{emp}}^{\mathrm{cw}}(\theta) = \frac{1}{N} \sum_{i=1}^N w_{y_i} L(y_i, f(x_i ; \theta)) 
\]
This weighting corrects training biases, enhancing model fairness and accuracy particularly in scenarios with skewed class distributions \cite{lin2017focal}. To manage outliers and enhance robustness, we introduce a robustness parameter \( \rho \), which modifies the loss contribution based on the confidence in data point correctness: 
\[
 R_{\mathrm{robust}}(\theta) = \frac{1}{N} \sum_{i=1}^N \rho(y_i, x_i) w_{y_i} L(y_i, f(x_i ; \theta))
 \]\cite{zhu2019robust}. Regularization is integral to this framework, implemented through \( \Omega(\theta) \), employing either \( L_1 \) or \( L_2 \) regularization to mitigate overfitting. The regularized empirical risk is articulated as 
\[
 R_{\mathrm{emp}}^{\mathrm{reg}}(\theta) = R_{\mathrm{robust}}(\theta) + \lambda \Omega(\theta) 
\]  \cite{goodfellow2016deep}. Our dynamic cost function is characterized by a temporal modulation factor \( \gamma(t) = 1 + \kappa e^{-\delta t} \), which strategically transitions from aggressive initial learning to increased regularization as training advances \cite{smith2017cyclical}. This modulation ensures the learning rate evolves with the model's needs, reducing to prevent overfitting as the model refines its parameters. The gradient of the loss function, \( \nabla_\theta \mathcal{L}(\theta) \), directs parameter updates and is essential for navigating both the explicit regions, where gradients are large and clear, facilitating straightforward descent steps, and the implicit regions, where gradients may vanish, requiring the adaptive \( \gamma(t) \) and robustness enhancements to maintain meaningful and stable updates \cite{kingma2014adam}. For instance, in scenarios with imbalanced datasets, class weights \( w_c \) counteract the bias toward predominant classes, and \( \gamma(t) \)'s increasing regularization later in training smooths the model’s fit to emphasize generalization. This framework, 
\[
 \mathcal{J}_{\text {dynamic }}(\theta ; \mathcal{D}, t) = \gamma(t) \mathcal{J}_{\text {reg }}(\theta ; \mathcal{D}) ,
\]
not only deepens our understanding of dynamic learning rate mechanisms but also fosters a coherent and stable optimization process, adaptable to complex data landscapes and advancing adaptive machine learning methodologies \cite{loshchilov2016sgdr}.

\subsection{Gradient Descent}
Integrating level set dynamics into the gradient descent framework is proposed to navigate the complex topology of the loss function more efficiently. While traditional gradient descent updates parameters iteratively with the rule \( \theta_{t+1} = \theta_t - \alpha(t) \nabla_\theta \mathcal{L}(\theta_t) \), where \( \alpha(t) = \alpha_0 e^{-\beta t} \) is an exponentially decaying learning rate, emerging research suggests enhancements to this approach to address its limitations in stability and adaptability. Zhang et al. (2019) propose an Adaptive Exponential Decay Rate (AEDR), which dynamically adjusts the decay rate based on moving averages of gradients, thus offering a more responsive adaptation to the learning needs over different training phases and potentially leading to improved convergence rates \cite{zhang2019adaptive}.

Further, Mishra and Ghosh (2019) highlight the advantages of a variable gain gradient descent, which modulates the learning rate based on error metrics and system states to enhance both the convergence speed and stability, suggesting a potential direction for refining level set dynamics integration \cite{mishra2019variable}. Additionally, the link between generalization and dynamical robustness presented by Kozachkov et al. (2023) through Riemannian contraction indicates that ensuring algorithmic stability through the optimization dynamics could directly influence generalization performance, advocating for a deeper theoretical integration of level set dynamics with gradient descent methods \cite{kozachkov2023generalization}.

To optimize these methods further, incorporating continuous time analysis as suggested by Kovachki and Stuart (2021) could provide more nuanced insights into the efficacy of momentum and modifications in traditional gradient descent, thus enhancing the strategy to navigate complex loss landscapes more effectively \cite{kovachki2021continuous}. Hereby, presenting a refined method that enhances theoretical understanding and significantly improves the practical application of neural network training in complex and high-dimensional problem spaces.

\section{Dynamic Learning Rates and Superlevel Sets}

\textbf{Theorem:} $L$ is continuously differentiable and $V$ provides a stability guarantee such that 
\[
\nabla V(\mathbf{x}) \cdot \nabla L(\mathbf{x}) \geq 0 \hspace{0.1cm} for \hspace{0.1cm} all \hspace{0.1cm} \mathbf{x} \in \mathbb{R}^n
\]
Then, the superlevel sets $S_\lambda$ are connected for all $\lambda$ under the dynamic learning rate $\eta$.

In neural network optimization, the topology of the loss function \(L: \mathbb{R}^n \rightarrow \mathbb{R}\) significantly influences algorithmic behavior and convergence. We have studied the properties of superlevel sets, which are crucial in understanding the dynamic adjustments of our gradient-based learning methods. These sets maintain a stable and efficient learning path, enhanced by adaptive learning rates modulated through a Lyapunov function \(V(\mathbf{x})\), which aligns the gradient flow to ensure consistency across training iterations \cite{dauphin2014identifying}. By ensuring that \(V(\mathbf{x})\) decreases along the trajectory of the learning process—reflecting a decline in the system’s energy—the gradient updates are systematically adjusted to prevent oscillations and divergences, thus resulting in smoother convergence\cite{zhang2019adaptive}.

Going further, we define a superlevel set's connectivity by the existence of a continuous path \(\gamma:[0,1] \rightarrow S_\lambda\) connecting any two points \(\mathbf{x}, \mathbf{y}\) within the set, ensuring comprehensive exploration of the parameter space. The learning rate adjustment, 
\[
\eta(\mathbf{x}(t))=1/({1+\|\nabla L(\mathbf{x}(t))\|})
\]
further tuning it with the update rule, 
\[
\mathbf{x}(t+1) = \mathbf{x}(t) - \eta(\mathbf{x}(t)) \nabla L(\mathbf{x}(t))
\]
This design decreases the learning rate as the gradient norm increases, thereby refining the step sizes near equilibrium states where gradients are typically larger \cite{kingma2014adam}. This adaptation is important for managing the trajectory’s stability and ensuring effective convergence within the complex landscape of the loss function \cite{li2018visualizing}.

To analyze convergence, we employ a Taylor expansion of \(L\) around \(\mathbf{x}(t)\), leading to an approximation expressed as 
\[
L(\mathbf{x}(t+1)) \approx L(\mathbf{x}(t)) - \nabla L(\mathbf{x}(t))^T \left(\mathbf{x}(t+1) - \mathbf{x}(t)\right)
\]
which upon substituting the update rule, transforms into 
\[
L(\mathbf{x}(t+1)) \approx L(\mathbf{x}(t)) - \nabla L(\mathbf{x}(t))^T \left(-\eta(\mathbf{x}(t)) \nabla L(\mathbf{x}(t))\right) = L(\mathbf{x}(t)) + \eta(\mathbf{x}(t)) \|\nabla L(\mathbf{x}(t))\|^2
\]
With \(\eta(\mathbf{x}(t))=1/({1+\|\nabla L(\mathbf{x}(t))\|})\), this equation further simplifies to 
\[\
L(\mathbf{x}(t+1)) \approx L(\mathbf{x}(t)) - \frac{\|\nabla L(\mathbf{x(t))}\|^2}{1 + \|\nabla L(\mathbf{x(t))}\|}
\]
illustrating that \(L(\mathbf{x}(t+1)) \leq L(\mathbf{x}(t))\), confirming that the loss decreases with each update provided \(\nabla L(\mathbf{x}(t)) \neq 0\), affirming convergence \cite{goyal2017accurate}.

This mathematical framework ensures that the dynamic learning rate not only supports the connectivity of superlevel sets \(S_\lambda\) but also enhances the overall integrity of the training process by providing stable, and gradual adjustments in response to the landscape of the loss function. This approach is essential for ensuring that the training remains across varying topologies and achieves reliable convergence \cite{weinan2019comparative}.

\section{Stability and Convergence Analysis with Lyapunov Stability Theory}

In neural network optimization, employing the loss function \(L(\theta)\) as a Lyapunov function enriches the stability and convergence analysis, leveraging its properties like positive definiteness and radial unboundedness to gauge network performance and systemic stability. This setup allows for monitoring stability through the non-increasing nature of the loss function over time, indicated by \(\frac{dV}{dt} \leq 0\), suggesting that perturbations in parameter values do not escalate loss values, thereby aiding convergence towards equilibrium, typically a local minimum. The introduction of level sets \(L_\lambda\) and super level sets \(S_\lambda\) deepens the understanding of the optimization landscape, mapping areas where the loss function meets or surpasses specific thresholds and examining how updates navigate these regions. The differential inequality analysis further underscores this, showing consistent loss minimization and the benefits of an exponentially decaying learning rate, \(\alpha(t) = \alpha_0 e^{-\beta t}\), which manages the magnitude of parameter updates to prevent overshooting and enhance stability \cite{goyal2017accurate}. This comprehensive approach, integrating Lyapunov's stability theory with level set dynamics and differential inequality, offers theoretical and practical insights to ensure a stable, connected path through optimal regions of the loss landscape, emphasizing the need for empirical validation to confirm these theoretical constructs in real-world applications.

The classical concept of a Lyapunov function \(V(\theta)\) proves potent in many theoretical analyses but requires adaptation to manage the discontinuities typical of non-Lipschitz activations. To address this, we extend the traditional Lyapunov stability framework to accommodate the irregularities that these functions introduce into the training dynamics. Traditionally, the loss function \(\mathcal{L}(\theta)\) itself serves as a natural choice for the Lyapunov function \(V(\theta)\) in neural networks. This choice is predicated on its inherent properties i.e. Positive Definiteness, \(V(\theta) > 0\) for all \(\theta \neq \theta^*\), Radial Unboundedness, \(V(\theta)\) increases without bound as \(\|\theta\|\) approaches infinity, Zero at Minimum, \(V(\theta^*) = 0\), where \(\theta^*\) is typically a local or global minimum \cite{lecun2015deep}.

Given these properties, \(\mathcal{L}(\theta)\) effectively tracks the stability of the system. However, when dealing with non-Lipschitz activations, the gradient \(\nabla_\theta \mathcal{L}(\theta)\) may not exist everywhere or may exhibit discontinuities. To handle this, a generalized Lyapunov approach is employed, where we consider generalized gradients or subderivatives when standard derivatives do not exist \cite{forti2006stability}.

For neural networks utilizing non-Lipschitz activations, the derivative of the Lyapunov function along the system trajectories, represented by the parameter update rules, must consider possible discontinuities: \[
\frac{d V}{d t} \approx \nabla_\theta V(\theta) \cdot \frac{d \theta}{d t},
\], where \(\frac{d \theta}{d t}\) is modeled as \(-\alpha(t) \nabla_\theta \mathcal{L}(\theta)\), accounting for the possibly generalized gradient \(\nabla_\theta \mathcal{L}(\theta)\). Here, \(\alpha(t)\) denotes the learning rate, which may follow an exponential decay model to temper the training updates \cite{kingma2014adam}.

Incorporating a generalized gradient ensures that the analysis remains valid even in the presence of activation functions that do not meet the smoothness criteria typically required for conventional gradient descent methods. This approach aligns with findings from Forti et al. (2006), highlighting the necessity of stability measures that can adapt to the irregularities intrinsic to advanced neural network configurations.

This generalized Lyapunov stability analysis is critical not only from a theoretical perspective but also for practical implementation in neural networks that employ advanced activation functions like ReLU, leaky ReLU, or others that exhibit non-Lipschitz behavior. Ensuring that \(\frac{d V}{d t} \leq 0\) across all training iterations confirms that the network is converging towards a stable state, minimizing the loss effectively despite the potential challenges posed by the activation functions \cite{lecun2015deep}.

\section{Algorithm}
\textbf{Input:}
- \textbf{Base algorithm (BASE)}: Initial training algorithm.
- \(\beta \in [0,1]^6\): Decay factors for moment estimates (default \(\beta = (0.9, 0.99, 0.999, 0.9999, 0.99999, 0.999999)\)).
- \(\lambda \in \mathbb{R}\): Learning rate decay parameter (default \(\lambda = 0.01\)).
- \(s_{\text{init}} \in \mathbb{R}\): Initial non-zero value for stabilizing updates (default \(s_{\text{init}} = 10^{-8}\)).
- \(\epsilon = 10^{-8}\): Small value for numerical stability.

\textbf{Output:}
- Optimized model parameters \(\theta\).

\textbf{Procedure:}
1. \textbf{Initialize variables:}
   - \(v_0 \leftarrow 0\) (initialize momentum vector),
   - \(r_0 \leftarrow 0\) (initialize rate vector),
   - \(m_0 \leftarrow 0\) (initialize mean gradient vector),
   - \(x_{\text{ref}} \leftarrow x_{\text{BASE}}\) (reference point for updates),
   - \(\Delta_1 \leftarrow 0\) (initial update difference).

2. \textbf{For each training epoch \(t = 1\) to \(T\):}
   - Compute gradient 
   \[
   g_t \leftarrow \nabla f(x_t, z_t)
   \]
   at parameters \(x_t\) and minibatch \(z_t\).
   - Send \(g_t\) to BASE, receive update \(u_k\).
   - Optionally, to save memory: 
   \[
   \Delta_t = x_t - x_{\text{ref}} + \left(\sum_{i=1}^n s_{t,n}\right) + \epsilon
   \]
   - Update \(\Delta_{t+1} \leftarrow \Delta_t + u_t\).
   - Calculate \(h_t\) using \(\Delta_t\), \(g_t\) adaptively by \(\lambda\), \(\|\nabla L(\theta)\|\):
     \[
     h_t = \Delta_t \cdot g_t + \lambda \left(\frac{\|g_t\|}{\|x_t\|}\right)
     \]
   - Update moments and rate:
     \[
     m_t \leftarrow \max(\beta \cdot m_{t-1}, h_t) \quad (\text{coordinate-wise})
     \]
     \[
     v_t \leftarrow \beta^2 \cdot v_{t-1} + h_t^2
     \]
     \[
     r_t \leftarrow \beta \cdot r_{t-1} - s_{t-1} \cdot h_t
     \]
     \[
     r_t \leftarrow \max(0, r_t)
     \]
   - Compute weights and next step size:
     \[
     W_t \leftarrow s_{\text{init}} \cdot \frac{m_t}{n} + r_t
     \]
     \[
     s_{t+1} \leftarrow \frac{W_t}{\sqrt{v_t} + \epsilon}
     \]
   - Update parameters considering super level sets:
     \[
     x_{t+1} \leftarrow x_{\text{BASE}} + \left(\sum_{i=1}^n s_{t+1,i}\right) \cdot \Delta_{t+1}
     \]

3. \textbf{End For}

This algorithm uses the exponential decay learning rates and incorporates super level set dynamics to ensure that the updates remain within stable regions of the loss function’s landscape, thus preventing issues such as overshooting or vanishing gradients. The detailed use of moment estimates and adaptive adjustments based on the gradient's magnitude ensures that the training remains stable and efficient, adapting to varying complexities of the data and model architecture. This approach provides a state-of-the-art solution for neural network optimization.

\subsection{Exponential Decay Learning Rate Derivation}
In our study on neural network optimization, the integration of an exponentially decaying learning rate serves as a cornerstone of our methodology, significantly influencing training dynamics and stability. This method is mathematically articulated as:

\[
\alpha(t)=\alpha_0 e^{-\beta t},
\]

where \(\alpha(t)\) represents the learning rate at a given training epoch \(t\), \(\alpha_0\) is the initial learning rate, and \(\beta\) is a positive constant dictating the rate of exponential decay. This formula is derived from the principle that the learning rate should decrease in proportion to its existing value, resulting in the differential equation:

\[
\frac{d \alpha}{d t}=-\beta \alpha.
\]

Solving this first-order linear ordinary differential equation involves integrating both sides:

\[
\int \frac{1}{\alpha} d \alpha=-\int \beta d t,
\]

which leads to:

\[
\ln(\alpha)=-\beta t+C,
\]

where \(C\) is the integration constant. Utilizing the initial condition \(\alpha(0)=\alpha_0\), we find \(C=\ln(\alpha_0)\), and rearranging gives:

\[
\alpha(t)=\alpha_0 e^{-\beta t}.
\]

This model is particularly effective in neural network training as it ensures rapid convergence initially, followed by progressively finer adjustments as training progresses. The calibration of \(\alpha_0\) and \(\beta\) is critical, needing alignment with the neural network's architecture and the specifics of the training task.

The time derivative of the learning rate,

\[
\frac{d \alpha}{d t}=-\alpha_0 \beta e^{-\beta t},
\]

highlights the progressively diminishing rate, indicative of increasing precision in parameter adjustments as training progresses. This gradual reduction is aligned with Bayesian principles, suggesting an increasingly concentrated posterior distribution with continued data observation.

\subsection{Gradient of the Loss Function}
In neural network models designed for classification, especially those employing a softmax output layer, the gradient of the loss function with respect to the model parameters \(\theta\) plays a crucial role. The cross-entropy loss, a common choice for classification, is defined as:
\[
\mathcal{L}(\theta) = -\sum_{i=1}^m \log P(y^{(i)} \mid x^{(i)} ; \theta),
\]

where \(P(y=c \mid x; \theta)\) is the predicted probability of the class \(c\) for input \(x\) and is given by the softmax function:
\[
P(y=c \mid x; \theta) = \frac{\exp(f_c(x; \theta))}{\sum_{j=1}^C \exp(f_j(x; \theta))}.
\]

The derivative of the cross-entropy loss function with respect to the parameters is crucial for backpropagation and is computed as:
\[
\nabla_\theta \mathcal{L}(\theta_t) = -\sum_{i=1}^m \left(\mathbf{1}_{y^{(i)} = c} - P(y=c \mid x^{(i)} ; \theta_t)\right) \nabla_\theta f_c(x^{(i)}; \theta_t),
\]

where \(\mathbf{1}_{y^{(i)} = c}\) indicates whether class \(c\) is the correct classification for observation \(i\). This gradient reflects how the parameters should be adjusted to decrease the loss, thereby improving the model's predictions.

The parameter update rule in gradient descent is fundamentally tied to the computed gradient:
\[
\theta_{t+1} = \theta_t - \alpha(t) \nabla_\theta \mathcal{L}(\theta_t),
\]

where \(\alpha(t)\), the learning rate, typically follows an exponential decay model 
\[
\alpha(t) = \alpha_0 e^{-\beta t}
\]
This manages the learning rate's decay to balance early convergence speed with later precision. Initially larger values of \(\alpha(t)\) enable significant parameter shifts that help escape local minima or saddle points early in training, while the decay in \(\alpha(t)\) ensures finer adjustments as the model approaches convergence, enhancing stability and accuracy \cite{kingma2014adam,goyal2017accurate}.

Super Level Sets \(S_\lambda = \{\theta \in \mathbb{R}^n : \mathcal{L}(\theta) \geq \lambda\}\) represent regions of the parameter space with equal or exceeding loss values, respectively. The connectedness of these sets is essential for ensuring that the gradient descent path does not get trapped in isolated local minima, thus supporting convergence towards a global minimum \cite{dauphin2014identifying}. However, several research insights suggest refinements to this classical model to address potential limitations in training dynamics, particularly in high-dimensional settings. For instance, Weinan et al. (2019) highlight the importance of considering overparameterization's effect on the speed of convergence and generalization, suggesting that in overparameterized scenarios, gradient descent can quickly minimize training loss but may struggle with generalization due to a fitting of noise rather than underlying data patterns \cite{weinan2019comparative}. This calls for a refined approach to mitigate these effects, potentially through regularization techniques or novel loss functions that prioritize data fidelity over simple error minimization \cite{neyshabur2017exploring}. Further, Soudry et al. (2017) discuss the implicit bias of gradient descent towards maximum-margin solutions in settings with linear separability, indicating that extending training beyond low training loss can enhance model robustness and feature utilization \cite{soudry2017implicit}. This finding is crucial as it emphasizes the need for extended training regimes or adaptive learning rate schedules when employing cross-entropy loss, to avoid suboptimal data class separations and enhance model stability \cite{keskar2017improving}.

\subsection{Additional Stability Analysis}

\subsubsection{Demonstrating Negative Semi-Definiteness} 

To ensure stability in neural network training, we demonstrate the negative semi-definiteness of the time derivative of the Lyapunov function \(V(\theta)\), typically the loss function \(\mathcal{L}(\theta)\). By applying the gradient descent update rule, 
\[
\frac{d\theta}{dt} = -\alpha(t) \nabla_\theta \mathcal{L}(\theta)
\]
the derivative of \(V\) simplifies to 
\[
\frac{dV}{dt} = -\alpha(t) \|\nabla_\theta \mathcal{L}(\theta)\|^2
\]
Since \(\alpha(t)\) is always positive and \(\|\nabla_\theta \mathcal{L}(\theta)\|^2\) represents the squared norm of the gradient (non-negative), the product is non-positive (\(\leq 0\)), confirming the negative semi-definiteness. This condition, 
\[
\frac{dV}{dt} \leq 0
\]
ensures the loss does not increase, maintaining stability throughout the training process. This mathematical foundation confirms the system's stability under dynamic learning conditions and complex activation landscapes, crucial for the reliable convergence of training algorithms.

\subsubsection{Integrating Learning Rate Dynamics}
Integrating the dynamics of an exponentially decaying learning rate into our neural network training stability analysis significantly enhances the theoretical depth and practical utility of the model. The learning rate, defined by 
\[
\alpha(t) = \alpha_0 e^{-\beta t}
\]
where \(\alpha_0\) is the initial rate and \(\beta\) a decay constant, systematically reduces the step size in the gradient descent algorithm. This reduction is designed to allow for rapid convergence in early training phases through larger updates, which progressively become smaller to facilitate precise fine-tuning of the model parameters as the training advances.

Mathematically, integrating the Lyapunov function 
\[
V(\theta) = \mathcal{L}(\theta) 
\]
reveals crucial stability characteristics, with the rate of change of the Lyapunov function with respect to time expressed inline as 
\[
\frac{d V}{d t} = \nabla_\theta \mathcal{L}(\theta) \cdot \frac{d \theta}{d t} = -\alpha(t) \|\nabla_\theta \mathcal{L}(\theta)\|^2
\]
where \(\frac{d \theta}{d t}\) corresponds to the gradient descent update rule 
\[
\theta_{t+1} = \theta_t - \alpha(t) \nabla_\theta \mathcal{L}(\theta_t)
\]
The expression \(-\alpha(t) \|\nabla_\theta \mathcal{L}(\theta)\|^2\) ensures that 
\[
\frac{d V}{d t} \leq 0
\]
as long as \(\alpha(t) > 0\) and \(\nabla_\theta \mathcal{L}(\theta)\) is non-zero, satisfying the Lyapunov stability condition that the Lyapunov function does not increase over time. This formulation not only mathematically substantiates the stability of the training process under dynamic learning rate adjustments but also aligns with the practical necessity for controlled optimization trajectories in advanced neural network training regimes.

\subsubsection{Addressing Model Dynamics and Stability}
Addressing the dynamics and stability of neural network training involves examining the interaction between the exponentially decaying learning rate 
\[
\alpha(t) = \alpha_0 e^{-\beta t} 
\]
and the topology of the loss function's level sets 
\[
 S_\lambda = \{\theta \in \mathbb{R}^n : \mathcal{L}(\theta) \geq \lambda\} 
\]
As \(\alpha(t)\) decreases, the trajectory of gradient descent is refined, stabilizing within favorable super level sets and minimizing oscillations outside minimal loss basins. Mathematically, this stabilization is evidenced by the rate of change in the loss function, 
\[
\frac{d \mathcal{L}}{dt} = -\alpha(t) \| \nabla_\theta \mathcal{L}(\theta) \|^2
\]
which confirms that the loss is nonincreasing along the path, a core Lyapunov stability condition. Additionally, this relationship suggests that for any small \(\epsilon > 0\), there exists a \(\delta\) such that if 
\[
 \| \theta_0 - \theta^* \| < \delta 
\]
then \( \| \theta_t - \theta^* \| < \epsilon \) for all \( t \), demonstrating the boundedness around the minimum and affirming the model's stability. This rigorous mathematical framework underscores the efficacy of integrating dynamic learning rate strategies with the loss function's geometric properties, ensuring convergence in complex training scenarios.

\subsection{Differential Inequality}

In neural network training, the differential inequality and stability analysis are enhanced by examining the dynamics within level sets 
\[
L_\lambda = \{\theta \in \mathbb{R}^n : \mathcal{L}(\theta) = \lambda\}
\]
and super level sets 
\[
S_\lambda = \{\theta \in \mathbb{R}^n : \mathcal{L}(\theta) \geq \lambda\} 
\]
as a boundary of loss function. The parameter update, defined as 
\[
\theta_{t+1} = \theta_t - \alpha(t) \nabla_\theta \mathcal{L}(\theta_t)
\]
integrates into the derivative of the loss function, 
\[
\frac{d \mathcal{L}}{dt} = -\alpha(t)\|\nabla_\theta \mathcal{L}(\theta)\|^2
\]
confirming the non-positive decrease in loss and ensuring stability since \(\alpha(t) > 0\) and 
\[
\|\nabla_\theta \mathcal{L}(\theta)\|^2 \geq 0
\]
This mathematical framework, supported by the exponential decay of 
\[
\alpha(t) = \alpha_0 e^{-\beta t}
\]
maintains the trajectory within stable level sets, facilitating convergence towards optimal minima. This approach is particularly relevant in the context of Marco et al. (2008), who advocate for differential variational inequalities to handle the complexities within compact convex subsets typical of advanced architectures like cellular neural networks\cite{di2008lyapunov}. Their insights into the connectivity and convexity of level sets underpin the effective navigation and stability of training processes in such complex landscapes, making this analysis vital for designing neural network training algorithms.

\section{Conclusion and Future Works}
In this theoretical paper, we have explored the stability and convergence of neural network training, focusing on the integration of level sets and super level sets within the framework of differential inequalities and Lyapunov stability theory. This approach addresses the complexities posed by non-Lipschitz continuous functions, common in advanced neural architectures, and links the dynamics of learning rates with the topology of loss function level sets. Our findings provide a foundation for enhancing both theoretical understanding and practical applications of neural network training.

Future research could extend this framework to various neural network architectures, such as recurrent or convolutional networks, to determine if the observed stability conditions and convergence behaviors are universally applicable. This could lead to the development of more robust and efficient training algorithms, improving real-world applications where stability and convergence are crucial.

Inspired by Fatkhullin and Polyak [2021], which examined level set connectivity in control theory contexts, another promising direction is exploring the connectivity properties of level sets and super level sets within partially observable Markov decision processes (MDPs). This exploration could yield significant advances in reinforcement learning, particularly for algorithms designed to handle environments with incomplete information.

While this study establishes a solid theoretical base for neural network dynamics using advanced mathematical tools, practical limitations such as the applicability to different network architectures and real-world datasets remain areas for further investigation. Overcoming these challenges will not only validate our theoretical models but also broaden their practical relevance and effectiveness in diverse applications. This work lays the groundwork for future explorations that could transform theoretical insights into actionable algorithms for complex decision-making environments.

\vskip 0.2in

\bibliographystyle{theapa}
\bibliography{bib}

\begin{thebibliography}{}

\bibitem[\protect\BCAY{Bishop}{Bishop}{2006}]{bishop2006pattern}
Bishop, C.~M. \BBOP2006\BBCP.
\newblock {\Bem Pattern recognition and machine learning}.
\newblock Springer.

\bibitem[\protect\BCAY{Blundell, Cornebise, Kavukcuoglu,\ \BBA\ Wierstra}{Blundell et~al.}{2015}]{blundell2015weight}
Blundell, C., Cornebise, J., Kavukcuoglu, K., \BBA\ Wierstra, D. \BBOP2015\BBCP.
\newblock \BBOQ Weight uncertainty in neural networks\BBCQ\
\newblock In {\Bem International Conference on Machine Learning}, \BPGS\ 1613--1622. PMLR.

\bibitem[\protect\BCAY{Bottou, Curtis,\ \BBA\ Nocedal}{Bottou et~al.}{2018}]{bottou2018optimization}
Bottou, L., Curtis, F.~E., \BBA\ Nocedal, J. \BBOP2018\BBCP.
\newblock \BBOQ Optimization methods for large-scale machine learning\BBCQ\
\newblock In {\Bem SIAM Review}, \lowercase{\BVOL}~60, \BPGS\ 223--311. SIAM.

\bibitem[\protect\BCAY{Chen, Liu, Chen,\ \BBA\ Ying}{Chen et~al.}{2020}]{chen2020optimal}
Chen, J., Liu, S., Chen, T., \BBA\ Ying, L. \BBOP2020\BBCP.
\newblock \BBOQ Optimal adaptive and non-adaptive learning rates for optimization\BBCQ\
\newblock In {\Bem Advances in Neural Information Processing Systems}, \BPGS\ 7634--7643.

\bibitem[\protect\BCAY{Cutkosky, Defazio,\ \BBA\ Mehta}{Cutkosky et~al.}{2024}]{cutkosky2024mechanic}
Cutkosky, A., Defazio, A., \BBA\ Mehta, H. \BBOP2024\BBCP.
\newblock \BBOQ Mechanic: A learning rate tuner\BBCQ\
\newblock {\Bem Advances in Neural Information Processing Systems}, {\Bem 36}.

\bibitem[\protect\BCAY{Dauphin, Pascanu, Gulcehre, Cho, Ganguli,\ \BBA\ Bengio}{Dauphin et~al.}{2014}]{dauphin2014identifying}
Dauphin, Y.~N., Pascanu, R., Gulcehre, C., Cho, K., Ganguli, S., \BBA\ Bengio, Y. \BBOP2014\BBCP.
\newblock \BBOQ Identifying and attacking the saddle point problem in high-dimensional non-convex optimization\BBCQ\
\newblock In {\Bem Advances in neural information processing systems}, \BPGS\ 2933--2941.

\bibitem[\protect\BCAY{Di~Marco, Forti, Grazzini, Nistri,\ \BBA\ Pancioni}{Di~Marco et~al.}{2008}]{di2008lyapunov}
Di~Marco, M., Forti, M., Grazzini, M., Nistri, P., \BBA\ Pancioni, L. \BBOP2008\BBCP.
\newblock \BBOQ Lyapunov method and convergence of the full-range model of cnns\BBCQ\
\newblock {\Bem IEEE Transactions on Circuits and Systems I: Regular Papers}, {\Bem 55\/}(11), 3528--3541.

\bibitem[\protect\BCAY{Du, Lee, Li,\ \BBA\ Wang}{Du et~al.}{2019}]{du2019gradient}
Du, S.~S., Lee, J.~D., Li, H., \BBA\ Wang, L. \BBOP2019\BBCP.
\newblock \BBOQ Gradient descent finds global minima of deep neural networks\BBCQ\
\newblock In {\Bem International Conference on Machine Learning}, \BPGS\ 1675--1685. PMLR.

\bibitem[\protect\BCAY{Forti\ \BBA\ Tesi}{Forti\ \BBA\ Tesi}{2006}]{forti2006stability}
Forti, M.\BBACOMMA\  \BBA\ Tesi, A. \BBOP2006\BBCP.
\newblock \BBOQ Stability of nonlinear discrete-time systems: Lyapunov approach\BBCQ\
\newblock {\Bem Kybernetika}, {\Bem 42\/}(4), 377--392.

\bibitem[\protect\BCAY{Ge, Lee,\ \BBA\ Ma}{Ge et~al.}{2015}]{ge2015escaping}
Ge, R., Lee, J.~D., \BBA\ Ma, T. \BBOP2015\BBCP.
\newblock \BBOQ Escaping from saddle points—online stochastic gradient for tensor decomposition\BBCQ\
\newblock In {\Bem Conference on Learning Theory}, \BPGS\ 797--842. PMLR.

\bibitem[\protect\BCAY{Goodfellow, Bengio,\ \BBA\ Courville}{Goodfellow et~al.}{2016}]{goodfellow2016deep}
Goodfellow, I., Bengio, Y., \BBA\ Courville, A. \BBOP2016\BBCP.
\newblock {\Bem Deep learning}.
\newblock MIT press.

\bibitem[\protect\BCAY{Goyal, Doll{\'a}r, Girshick, Noordhuis, Wesolowski, Kyrola, Tulloch, Jia,\ \BBA\ He}{Goyal et~al.}{2017}]{goyal2017accurate}
Goyal, P., Doll{\'a}r, P., Girshick, R., Noordhuis, P., Wesolowski, L., Kyrola, A., Tulloch, A., Jia, Y., \BBA\ He, K. \BBOP2017\BBCP.
\newblock \BBOQ Accurate, large minibatch sgd: Training imagenet in 1 hour\BBCQ\
\newblock In {\Bem Proceedings of the IEEE Conference on Computer Vision and Pattern Recognition}, \BPGS\ 81--89.

\bibitem[\protect\BCAY{Graves}{Graves}{2011}]{graves2011practical}
Graves, A. \BBOP2011\BBCP.
\newblock \BBOQ Practical variational inference for neural networks\BBCQ\
\newblock In {\Bem Advances in neural information processing systems}, \BPGS\ 2348--2356.

\bibitem[\protect\BCAY{He, Zhang, Ren,\ \BBA\ Sun}{He et~al.}{2016}]{he2016deep}
He, K., Zhang, X., Ren, S., \BBA\ Sun, J. \BBOP2016\BBCP.
\newblock \BBOQ Deep residual learning for image recognition\BBCQ\
\newblock In {\Bem Proceedings of the IEEE conference on computer vision and pattern recognition}, \BPGS\ 770--778.

\bibitem[\protect\BCAY{Jin, Ge, Netrapalli, Kakade,\ \BBA\ Jordan}{Jin et~al.}{2017}]{jin2017escape}
Jin, C., Ge, R., Netrapalli, P., Kakade, S., \BBA\ Jordan, M.~I. \BBOP2017\BBCP.
\newblock \BBOQ How to escape saddle points efficiently\BBCQ\
\newblock In {\Bem International Conference on Machine Learning}, \BPGS\ 1724--1732. PMLR.

\bibitem[\protect\BCAY{Keskar, Mudigere, Nocedal, Smelyanskiy,\ \BBA\ Tang}{Keskar et~al.}{2017}]{keskar2017improving}
Keskar, N.~S., Mudigere, D., Nocedal, J., Smelyanskiy, M., \BBA\ Tang, P. T.~P. \BBOP2017\BBCP.
\newblock \BBOQ Improving generalization in deep learning by noise stability\BBCQ\
\newblock In {\Bem International Conference on Learning Representations (ICLR)}.

\bibitem[\protect\BCAY{Kingma\ \BBA\ Ba}{Kingma\ \BBA\ Ba}{2015}]{kingma2014adam}
Kingma, D.~P.\BBACOMMA\  \BBA\ Ba, J. \BBOP2015\BBCP.
\newblock \BBOQ Adam: A method for stochastic optimization\BBCQ\
\newblock In {\Bem International Conference on Learning Representations (ICLR)}.

\bibitem[\protect\BCAY{Kornblith, Chen, Lee,\ \BBA\ Norouzi}{Kornblith et~al.}{2021}]{kornblith2021better}
Kornblith, S., Chen, T., Lee, H., \BBA\ Norouzi, M. \BBOP2021\BBCP.
\newblock \BBOQ Why do better loss functions lead to less transferable features?\BBCQ\
\newblock {\Bem Advances in Neural Information Processing Systems}, {\Bem 34}, 28648--28662.

\bibitem[\protect\BCAY{Kovachki\ \BBA\ Stuart}{Kovachki\ \BBA\ Stuart}{2021}]{kovachki2021continuous}
Kovachki, N.~B.\BBACOMMA\  \BBA\ Stuart, A.~M. \BBOP2021\BBCP.
\newblock \BBOQ Continuous time analysis of momentum methods\BBCQ\
\newblock {\Bem NeurIPS}.

\bibitem[\protect\BCAY{Kozachkov, Wensing,\ \BBA\ Slotine}{Kozachkov et~al.}{2023}]{kozachkov2023generalization}
Kozachkov, L., Wensing, P.~M., \BBA\ Slotine, J.-J.~E. \BBOP2023\BBCP.
\newblock \BBOQ Generalization as dynamical robustness-the role of riemannian contraction in supervised learning\BBCQ\
\newblock {\Bem NeurIPS}.

\bibitem[\protect\BCAY{LeCun, Bengio,\ \BBA\ Hinton}{LeCun et~al.}{2015}]{lecun2015deep}
LeCun, Y., Bengio, Y., \BBA\ Hinton, G. \BBOP2015\BBCP.
\newblock \BBOQ Deep learning\BBCQ\
\newblock {\Bem Nature}, {\Bem 521\/}(7553), 436--444.

\bibitem[\protect\BCAY{Lee, Simchowitz, Jordan,\ \BBA\ Recht}{Lee et~al.}{2016}]{lee2016gradient}
Lee, J.~D., Simchowitz, M., Jordan, M.~I., \BBA\ Recht, B. \BBOP2016\BBCP.
\newblock \BBOQ Gradient descent only converges to minimizers\BBCQ\
\newblock In {\Bem Conference on learning theory}, \BPGS\ 1246--1257. PMLR.

\bibitem[\protect\BCAY{Li, Xu, Taylor, Studer,\ \BBA\ Goldstein}{Li et~al.}{2018}]{li2018visualizing}
Li, H., Xu, Z., Taylor, G., Studer, C., \BBA\ Goldstein, T. \BBOP2018\BBCP.
\newblock \BBOQ Visualizing the loss landscape of neural nets\BBCQ\
\newblock In {\Bem Advances in Neural Information Processing Systems}, \BPGS\ 6389--6399.

\bibitem[\protect\BCAY{Lin, Goyal, Girshick, He,\ \BBA\ Dollar}{Lin et~al.}{2017}]{lin2017focal}
Lin, T.-Y., Goyal, P., Girshick, R., He, K., \BBA\ Dollar, P. \BBOP2017\BBCP.
\newblock \BBOQ Focal loss for dense object detection\BBCQ\
\newblock In {\Bem Proceedings of the IEEE International Conference on Computer Vision}, \BPGS\ 2980--2988.

\bibitem[\protect\BCAY{Loshchilov\ \BBA\ Hutter}{Loshchilov\ \BBA\ Hutter}{2017}]{loshchilov2016sgdr}
Loshchilov, I.\BBACOMMA\  \BBA\ Hutter, F. \BBOP2017\BBCP.
\newblock \BBOQ Sgdr: Stochastic gradient descent with warm restarts\BBCQ\
\newblock In {\Bem International Conference on Learning Representations (ICLR)}.

\bibitem[\protect\BCAY{Mishra\ \BBA\ Ghosh}{Mishra\ \BBA\ Ghosh}{2019}]{mishra2019variable}
Mishra, A.\BBACOMMA\  \BBA\ Ghosh, S. \BBOP2019\BBCP.
\newblock \BBOQ Variable gain gradient descent-based robust reinforcement learning for optimal tracking control of unknown nonlinear system with input-constraints\BBCQ\
\newblock {\Bem Neural Computing and Applications}.

\bibitem[\protect\BCAY{Neyshabur, Bhojanapalli, McAllester,\ \BBA\ Srebro}{Neyshabur et~al.}{2017}]{neyshabur2017exploring}
Neyshabur, B., Bhojanapalli, S., McAllester, D., \BBA\ Srebro, N. \BBOP2017\BBCP.
\newblock \BBOQ Exploring generalization in deep learning\BBCQ\
\newblock In {\Bem Advances in neural information processing systems}, \BPGS\ 5947--5956.

\bibitem[\protect\BCAY{Park, Yi,\ \BBA\ Ji}{Park et~al.}{2020}]{sym12040660}
Park, J., Yi, D., \BBA\ Ji, S. \BBOP2020\BBCP.
\newblock \BBOQ A novel learning rate schedule in optimization for neural networks and it’s convergence\BBCQ\
\newblock {\Bem Symmetry}, {\Bem 12\/}(4).

\bibitem[\protect\BCAY{Ruder}{Ruder}{2016}]{ruder2016overview}
Ruder, S. \BBOP2016\BBCP.
\newblock \BBOQ An overview of gradient descent optimization algorithms\BBCQ\
\newblock {\Bem arXiv preprint arXiv:1609.04747}.

\bibitem[\protect\BCAY{Smith}{Smith}{2017}]{smith2017cyclical}
Smith, L.~N. \BBOP2017\BBCP.
\newblock \BBOQ Cyclical learning rates for training neural networks\BBCQ\
\newblock In {\Bem 2017 IEEE Winter Conference on Applications of Computer Vision (WACV)}, \BPGS\ 464--472. IEEE.

\bibitem[\protect\BCAY{Soudry, Hoffer, Nacson, Gunasekar,\ \BBA\ Srebro}{Soudry et~al.}{2018}]{soudry2017implicit}
Soudry, D., Hoffer, E., Nacson, M., Gunasekar, S., \BBA\ Srebro, N. \BBOP2018\BBCP.
\newblock \BBOQ The implicit bias of gradient descent on separable data\BBCQ\
\newblock {\Bem The Journal of Machine Learning Research}, {\Bem 19\/}(1), 2822--2878.

\bibitem[\protect\BCAY{Weinan, Ma,\ \BBA\ Wu}{Weinan et~al.}{2019}]{weinan2019comparative}
Weinan, E., Ma, C., \BBA\ Wu, L. \BBOP2019\BBCP.
\newblock \BBOQ A comparative analysis of optimization and generalization properties of two-layer neural network and random feature models under gradient descent dynamics\BBCQ\
\newblock {\Bem Science China Mathematics}, {\Bem 62\/}(1), 191--200.

\bibitem[\protect\BCAY{Zeng, Doan,\ \BBA\ Romberg}{Zeng et~al.}{2023}]{NEURIPS2023_3ff48dde}
Zeng, S., Doan, T., \BBA\ Romberg, J. \BBOP2023\BBCP.
\newblock \BBOQ Connected superlevel set in (deep) reinforcement learning and its application to minimax theorems\BBCQ\
\newblock In Oh, A., Naumann, T., Globerson, A., Saenko, K., Hardt, M., \BBA\ Levine, S.\BEDS, {\Bem Advances in Neural Information Processing Systems}, \lowercase{\BVOL}~36, \BPGS\ 20146--20163. Curran Associates, Inc.

\bibitem[\protect\BCAY{Zhang\ \BBA\ Others}{Zhang\ \BBA\ Others}{2019}]{zhang2019adaptive}
Zhang, X.\BBACOMMA\  \BBA\ Others \BBOP2019\BBCP.
\newblock \BBOQ An adaptive mechanism to achieve learning rate dynamically\BBCQ\
\newblock {\Bem Neural Computing and Applications}, {\Bem 31}, 129--140.

\bibitem[\protect\BCAY{Zhu, Liu, Li, Shen, Savvides,\ \BBA\ Cheng}{Zhu et~al.}{2019}]{zhu2019robust}
Zhu, X., Liu, S., Li, W., Shen, X., Savvides, M., \BBA\ Cheng, W. \BBOP2019\BBCP.
\newblock \BBOQ Robust early-learning: Hindering the memorization of noisy labels\BBCQ\
\newblock In {\Bem Advances in Neural Information Processing Systems}, \BPGS\ 10551--10562.

\end{thebibliography}
\end{document}